\documentclass[11pt,a4paper]{article}
\usepackage[hyperref]{acl2021}
\usepackage{times}
\usepackage{latexsym}
\usepackage{bm}
\usepackage{amsmath}
\usepackage{amssymb}
\usepackage{amsfonts}
\usepackage{booktabs}
\usepackage{color, colortbl}
\usepackage{multirow}
\usepackage{graphicx}
\usepackage{tikz}
\usepackage{enumitem}
\usetikzlibrary{shapes.geometric}
\usetikzlibrary{arrows.meta,arrows}

\usepackage{bbm}
\usepackage{arydshln}

\usepackage{microtype}
\usepackage{pgfplots}
\pgfplotsset{compat=1.17}

\usepackage{soul}
\usepackage{color}
\usepackage{dcolumn}
\definecolor{bittersweet}{HTML}{ff9292}
\definecolor{aureolin}{HTML}{fdffbc}
\definecolor{blue-green}{HTML}{98ded9}
\definecolor{caribbeangreen}{HTML}{e9b0df}

\DeclareRobustCommand{\hlred}[1]{{\sethlcolor{bittersweet}\hl{#1}}}
\DeclareRobustCommand{\hlblue}[1]{{\sethlcolor{blue-green}\hl{#1}}}
\DeclareRobustCommand{\hlyellow}[1]{{\sethlcolor{aureolin}\hl{#1}}}
\DeclareRobustCommand{\hlgreen}[1]{{\sethlcolor{caribbeangreen}\hl{#1}}}

\aclfinalcopy 
\def\aclpaperid{3566} 

\DeclareMathOperator*{\argmax}{\arg\!\max}

\newcommand{\method}{\textsc{DYPLOC}}

\title{\textsc{DYPLOC}: Dynamic Planning of Content Using Mixed Language Models \\for Text Generation}

\author{
Xinyu Hua$^{1}$ \quad Ashwin Sreevatsa$^{2}$ \quad Lu Wang$^{2}$ \\
$^{1}$Khoury College of Computer Sciences, Northeastern University, Boston, MA \\
$^{2}$Computer Science and Engineering, University of Michigan, Ann Arbor, MI \\
$^{1}${\tt hua.x@northeastern.edu} \\
$^{2}${\tt \{asreeva, wangluxy\}@umich.edu} \\
}

\date{}

\begin{document}
\maketitle
\begin{abstract}
We study the task of long-form opinion text generation, which faces at least two distinct challenges. First, existing neural generation models fall short of coherence, thus requiring efficient content planning. Second, diverse types of information are needed to guide the generator to cover both subjective and objective content. 
To this end, we propose \textsc{DYPLOC}, a generation framework that conducts dynamic planning of content while generating the output based on a novel design of mixed language models. 
To enrich the generation with diverse content, we further propose to use large pre-trained models to predict relevant concepts and to generate claims.
We experiment with two challenging tasks on newly collected datasets: (1) argument generation with Reddit ChangeMyView, and (2) writing articles using New York Times' Opinion section. 
Automatic evaluation shows that our model significantly outperforms competitive comparisons. 
Human judges further confirm that our generations are more coherent with richer content.
\end{abstract}

\section{Introduction}
\label{sec:intro}

Opinion articles serve as an important media to convey the authors' values, beliefs, and stances on important societal issues. Automatically generating long-form opinion articles has the potential of facilitating various tasks, such as essay writing and speech drafting, and it is the focus of this work. Though opinion generation has been investigated for constructing arguments~\cite{hua-wang-2018-neural}, writing reviews~\cite{ni2018personalized}, and producing emotional dialogue responses~\cite{song2019generating}, those outputs are relatively short. While impressive progress in generation has been achieved by using large pre-trained Transformers~\cite{radford2019language,lewis2019bart}, directly adopting them for long-form opinion text generation poses distinct challenges. 

\begin{figure}[t]
    \hspace{-1mm}
    \includegraphics[width=75mm]{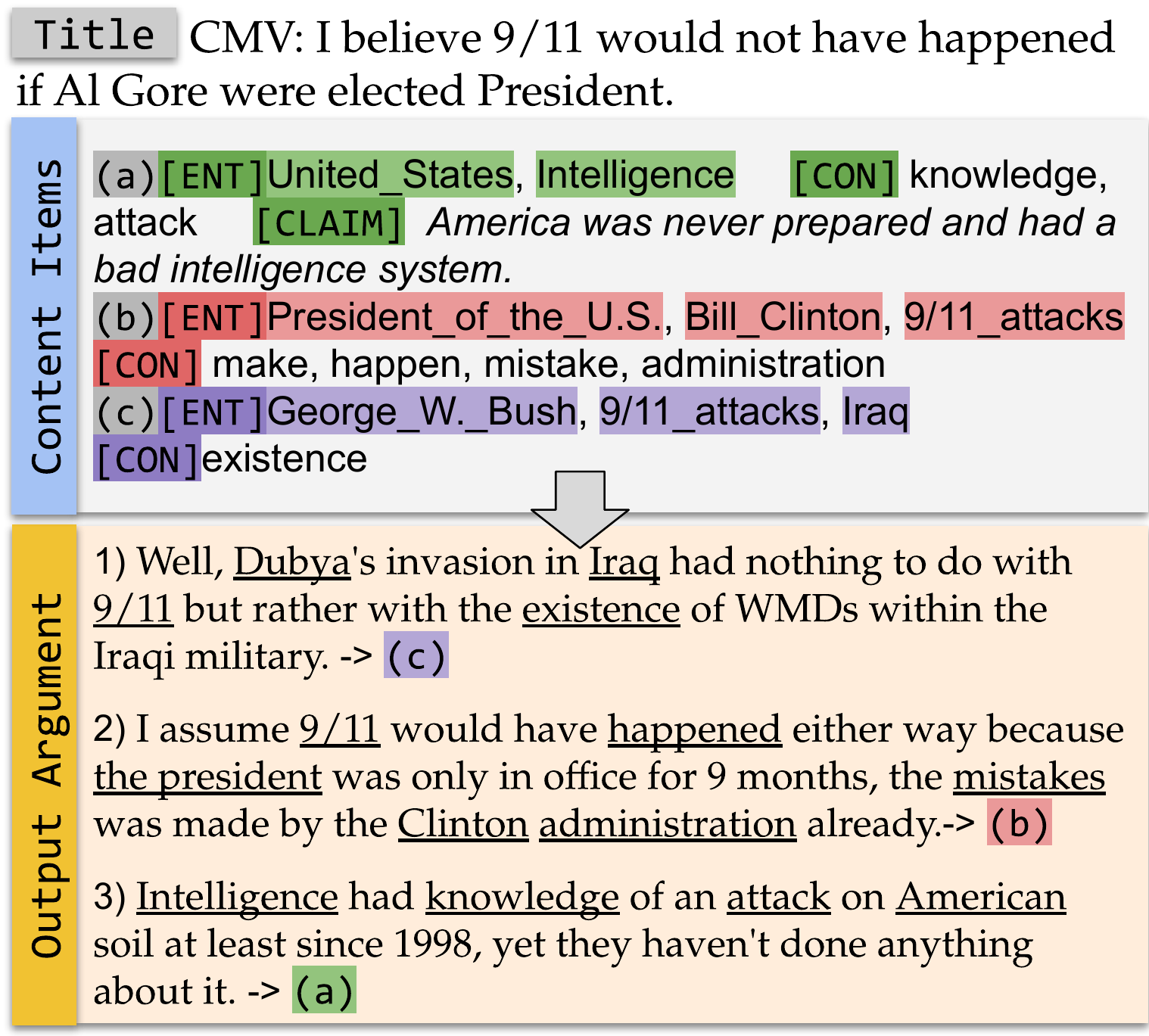}
    \caption{
Sample counter-argument on Reddit ChangeMyView. 
Our generator considers an input containing (1) a title and (2) an unordered set of content items. 
Each content item consists of {elements} of an \textit{entity set} \texttt{[ENT]}, a \textit{concept set} \texttt{[CON]}, and an optional one-sentence \textit{claim} \texttt{[CLAIM]}. 
Each output token is generated by conditioning on all content items, and the best aligned ones (learned by our model) are highlighted in corresponding colors. 
We also underline words that reflect the input concepts and entities. 
}
\label{fig:intro-example}

\end{figure}

First, large models still fall short of producing coherent text due to the \textit{lack of efficient content control and planning}~\cite{ko-li-2020-assessing,wu2020controllable,tan-etal-2021-progressive}.
A common solution is to use concatenated phrases or semantic representations to guide the generation process~\cite{yao2019plan,harkous-etal-2020-text,ribeiro2020investigating,goldfarb-tarrant-etal-2020-content}, where content planning, including both content selection and ordering, is expected to be learned by attention mechanisms. However, attentions have only achieved limited improvements.
Recent work also explores training a separate planning module to produce sorted content, which is then fed into a generator~\cite{fan-etal-2019-strategies,hua-wang-2020-pair,goldfarb-tarrant-etal-2020-content}. Nonetheless, this strategy results in a disconnection between planning and realization, and the output is not guaranteed to respect the planning results~\cite{castro-ferreira-etal-2019-neural,prabhumoye-etal-2020-exploring}. 

The second challenge for opinion generation resides in the diversity of information that is needed to produce an output with consistent stances and supported by pertinent facts. Though large models memorize significant amounts of knowledge, they cannot retrieve and operate with them precisely~\cite{lewis2020retrieval}. Due to the argumentative nature of opinion text, simply including knowledge bases~\cite{guan-etal-2020-knowledge,zhou2020improving} is insufficient to uphold the desired quality, as it requires the combination of subjective claims and objective evidence as supports. 

To this end, we propose a novel generation framework, \textsc{\textbf{DYPLOC}} (\underline{dy}namic \underline{pl}anning \underline{o}f \underline{c}ontent), to conduct content selection and ordering as text is produced.\footnote{Data and code are available at: \url{xinyuhua.github.io/Resources/acl21/}.}
Concretely, given a set of unordered content items, as displayed in Figure~\ref{fig:intro-example}, we design mixed language models, with each implemented as a sequence-to-sequence model to encode one item and the input statement. 
At each decoding step, our system selects which items to reflect, and predicts a word based on probabilities marginalized over all language models. 
Crucially, our end-to-end trained framework (1) enables the generator to access multiple content items at all times and select content based on what has been generated so far, (2) can be directly built on large pre-trained Transformers, e.g., BART~\cite{lewis2019bart}, with planning and generation modules jointly trained, and (3) outputs learned content selection scores to provide an interface for system decision interpretation. 

Furthermore, to ensure that our framework can be applied to a broad range of generation tasks, we design content items to cover three critical elements: \texttt{entities} and \texttt{concepts} that are central to many generation applications, and \texttt{claims} that are building blocks for opinion text. We show an example for counter-argument generation in Figure~\ref{fig:intro-example}. 
Importantly, we employ BART to predict additional relevant concepts, derived from ConceptNet~\cite{speer2017conceptnet}, and generate claims, as central propositions, to enrich the generated text with both objective and subjective content.

For experiments, we collect two datasets: (1) posts from Reddit ChangeMyView for argument generation, and (2) articles from the New York Times Opinion section~\cite{sandhaus2008new} for opinion article writing. 
Our proposed framework outperforms competitive comparisons, such as fine-tuning BART with the same content items, based on automatic metrics of BLEU, ROUGE, and METEOR. 
Human assessment further confirms that our system outputs have richer content and are more coherent in both tasks.

Our main contributions are summarized as below: 

$\bullet$ We present a dynamic content planning generation framework, which is directly built on top of BART. Our design of mixed language models overcomes the lack of control by existing models that use implicit planning with attentions or hard copying. 

$\bullet$ We propose content plan augmentation by automatically generating relevant concepts and claims.

$\bullet$ We construct two opinion text generation datasets with content plans that capture prominent entities and concepts.


\section{Related Work}
\label{sec:related}
\begin{figure*}[t]
    \centering
    \includegraphics[height=38mm]{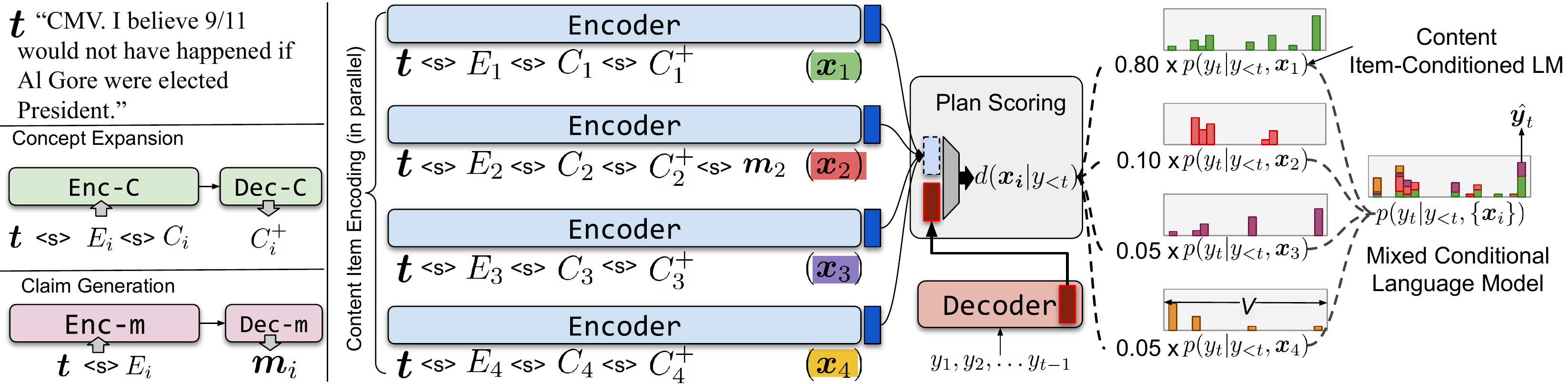}
    \vspace{-3mm}
    \caption{\fontsize{10}{12}\selectfont 
    Our proposed text generation framework, DYPLOC. 
    [Left] For each input content item (a title $\bm{t}$, an entity set ${E}_i$, and a core concept set ${C}_i$), we first expand it with more relevant concepts, i.e., ${C}^+_i$. 
    For sentences to be realized as claims, we employ a separate generator to produce one draft claim, $\bm{m}_i$. 
    [Right] The augmented content items, denoted as $\{\bm{x}_i\}$, are encoded in parallel. 
    At each decoding step, a plan scoring network estimates a distribution $d(\bm{x}_i|y_{<t})$ for all content items and decides on relevant content. A word is predicted based on probabilities marginalized over all content item-conditioned language models, i.e., $p(y_t|y_{<t}, \bm{x}_i)$ for the $i$-th model. 
    }
    \label{fig:model-overview}
\end{figure*}

\noindent \textbf{Neural Generation with Planning.} 
Text planning is seen as a crucial step to guide the generation of high-quality, well-organized natural language text~\cite{mckeown1992text,reiter2000building}. Incorporating planning modules to neural text generator has attracted significant research interests~\cite{shen-etal-2019-select,moryossef2019step,puduppully-etal-2019-data}, which proves to be especially beneficial for long-form output~\cite{fan-etal-2019-strategies,hua-wang-2019-sentence}. 
More recently, large pre-trained Transformers have established new state-of-the-arts for a wide range of text generation tasks~\cite{lewis2019bart,roller2020recipes,kale-rastogi-2020-text}. But it is non-trivial to integrate planning modules into them. 
Existing approaches resort to decoupling planning and decoding stages~\cite{hua-wang-2020-pair, kedzie-mckeown-2020-controllable}, which inevitably increases system complexities and potentially introduces cascading errors. 

We take inspiration from the retrieval-augmented generation framework~\cite{lewis2020retrieval}, which is designed to incorporate relevant documents for question answering. Our adaptation uses a trainable plan scoring module to reflect content selection and ordering, which is more suitable for long text generation and offers better interpretability. 
Concurrent work by \newcite{zhang2021joint} presents a mixture-of-expert decoder to tackle knowledge-grounded generation. However, their score distribution for language models is fixed across all decoding steps, whereas ours is updated as generation progresses and can better reflect the dynamic nature of content planning.

\paragraph{Controllable Text Generation.}
Another related line of research investigates the controllability of generation models~\cite{wiseman-etal-2017-challenges}, including conditioning over keywords~\cite{keskar2019ctrl,hua-wang-2020-pair,xu-etal-2020-megatron}, syntactic structures~\cite{casas-etal-2020-syntax,goyal-durrett-2020-neural}, or semantic representations~\cite{wen-etal-2015-semantically,elder-etal-2018-e2e}. 
Our work differs from all previous methods as we combine different types of content, covering both objective and subjective information, and attain fine-grained sentence-level control using a novel design of mixed conditional language models.

\smallskip
\noindent \textbf{Opinion Text Generation.} 
Our model tackles opinion articles, which differs from traditional text generation systems that mostly concern fact-based generations~\cite{gardent-etal-2017-webnlg,novikova-etal-2017-e2e,puduppully-etal-2019-data}. 
An extensive body of work has studied summarizing~\cite{wang-ling-2016-neural,suhara-etal-2020-opiniondigest,brazinskas-etal-2020-unsupervised} or generating~\cite{ni2018personalized,li-etal-2019-generating} reviews and building dialogue systems enhanced with emotions~\cite{li-etal-2016-persona,song2019generating}. 
More recently, developments are made in generating argumentative text~\cite{el-baff-etal-2019-computational,hidey-mckeown-2019-fixed}, which primarily focus on constructing single sentence claims on a limited number of topics. 
In comparison, our model can handle substantially longer output with improved quality.

\section{Model}
\label{sec:model}
\paragraph{Task Formulation.}
Our opinion text generation framework takes as input a set of \textbf{content items}. 
Each content item consists of a title $\bm{t}$, a set of \textbf{entities} $E_i$~\footnote{Note that $i$ distinguishes the items. Their order is random.}, 
such as \{\texttt{United\_States}, \texttt{9/11\_attacks}\}, and a set of \textbf{core concepts} $C_i$, such as \{\textit{attack}, \textit{knowledge}\}, that are often abstract notions. 
Our model first expands $C_i$ by predicting additional \textbf{relevant concepts} $C^+_i$ and optionally generates a pertinent \textbf{claim} $\bm{m}_i$, and then outputs the final text with multiple sentences as $\bm{y}=\{y_t\}$, to faithfully reflect the content items with a coherent structure. An overview of our system is illustrated in Figure~\ref{fig:model-overview}.

Below we first describe the content item augmentation methods (\S~\ref{sec:encoding}), followed by our generator with mixed language models that condition on expanded content items (\S~\ref{sec:mixed_cond}). 

\subsection{Content Item Augmentation}
\label{sec:encoding}

\paragraph{Concept Expansion.} 
With limited number of entities and concepts as input, generation systems are often incapable of producing long text with rich content, resulting in hallucination~\cite{wiseman-etal-2017-challenges,tian2019sticking}. 
Therefore, from the often-abstract core concepts, we aim to predict more specific concepts that are also relevant to the given title. For instance, as displayed in Figure~\ref{fig:intro-example}, for core concepts \{\textit{make}, \textit{happen}\} and entities \{\texttt{Bill\_Clinton}, \texttt{9/11\_attacks}\}, we grow the input with more concrete concepts of \{\textit{mistake}, \textit{administration}\}. 

We thus consider a concept expansion module $g (\cdot)$, which predicts additional relevant concepts, denoted as $C^+_i$, by conditioning on the original content item:

{\fontsize{10}{11}\selectfont
\setlength{\abovedisplayskip}{2pt}
\begin{align}
    C^+_i = g(\bm{t}, E_i, C_i) \label{eq:concept-gen}
\end{align}
}

While $g (\cdot)$ can be any conditional predictor, our experiment shows that fine-tuned BART model performs best on our tasks, where it generates $C^+_i$ word-by-word by consuming the content item.\footnote{We also exploited a model that uses the structure of knowledge bases, e.g., ConceptNet, for learning to expand concepts, but it yields lower precision and recall than fine-tuning BART does.} Training data construction is described in \S~\ref{sec:plan_data}.

\smallskip
\noindent\textbf{Claim Generation.} 
As discussed in \S~\ref{sec:intro}, opinion text generation should be controlled with consistent propositions, which cannot be effectively expressed by disconnected concepts. Therefore, we argue that natural languages are more suitable for delivering central claims, since they better encode stylistic languages, e.g., persuasion strategies. 

Concretely, we fine-tune another BART model by taking in the title $\bm{t}$ and the entities $E_i$, which then produces a claim with nucleus sampling for decoding~\cite{Holtzman2020The}. In this work, we assume the subset of content items that can be used to generate claims is known. Possible future work includes predicting such subsets and filtering claims with quality measurement. 

\subsection{Content Realization via Mixed Conditioning}
\label{sec:mixed_cond}

After obtaining the augmented content items, we leverage the BART model to encode each of them as a sequence, as illustrated in Figure~\ref{fig:model-overview}. Segmenter \texttt{<s>} is added to indicate the change of elements in a content item. 
Our encoders run over all items $\{\bm{x}_i\}$ in parallel, from which we extract content item representations $\{\bm{h}_i\}$, based on the last layer's hidden states of the first token. 

The standard sequence-to-sequence (seq2seq) framework models output probabilities by taking a single sequence as input. It is challenging to extend seq2seq to consider multiple sequences simultaneously, and conduct content planning concurrently. 
Therefore, we introduce \textbf{a plan scoring network}, $d(\bm{x}_i|y_{<t})$, which learns to dynamically select and order content based on what has been produced previously while generating the outputs. 
As outlined in Figure~\ref{fig:model-overview}, our generator is informed of all content items during generation. At each decoding step $t$, the probabilities of output words are estimated as a weighted sum of all content item-conditioned language models as follows:

{\fontsize{10}{11}\selectfont
\setlength{\abovedisplayskip}{2pt}
\begin{align}
    & p(y_{t}|y_{<t}) = \sum_{i} d(\bm{x}_i|y_{<t})p(y_t|y_{<t}, \bm{x}_i) \label{eq:marginalization} \\
    & d(\bm{x}_i|y_{<t}) = \text{softmax}_i(e_{it}) \label{eq:scoring}
\end{align}
}
where $p(y_t|y_{<t}, \bm{x}_i)$ corresponds to the $i$-th language model with $\bm{x}_i$ as the input. 
Crucially, $d(\bm{x}_i|y_{<t})$ determines the importance of $\bm{x}_i$ when generating token $y_t$ and thus achieves the effect of content planning. We design a two-layer feed-forward network to estimate $e_{it}$:

{\fontsize{10}{11}\selectfont
\setlength{\abovedisplayskip}{2pt}
\begin{align}
    & e_{it} = \bm{W}_o\tanh(\bm{W}_d[\bm{h}_i; \bm{s}_t]) \label{eq:score_pred}
\end{align}
}
where $\bm{h}_i$ denotes the representation of content item $\bm{x}_i$, $\bm{s}_t$ is the decoder state, and $\bm{W}_o$ and $\bm{W}_d$ are learnable parameters. 
Although mixed language models have been used by \newcite{lewis2020retrieval} to include retrieved documents for question answering, their relevance scores are given by external retrieval models, whereas our plan scorer $d(\bm{x}_i|y_{<t})$ is learned together with the generator. 

\smallskip
\noindent\textbf{Training and Decoding.} 
Our model is end-to-end trained with both the standard cross-entropy loss $\mathcal{L}_{gen}$ over the tokens in the target generations and a separate loss $\mathcal{L}_{plan}$ for learning $d(\bm{x}_i|y_{<t})$:

{\fontsize{10}{11}\selectfont
\setlength{\abovedisplayskip}{2pt}
\begin{align}
    &\mathcal{L}(\theta) = \mathcal{L}_{gen}(\theta) + \mathcal{L}_{plan}(\theta) 
\end{align}
}

To create labels for $\mathcal{L}_{plan}$, we leverage the correspondence between content items and target tokens, i.e., $d(\bm{x}_i|y_{<t})$ is optimized to approach $1$ if $y_i$ is in the sentence that derives $\bm{x}_i$, otherwise $0$.\footnote{We also experimented with a training objective consisting of the generation loss only, but the performance degraded significantly. Future directions include removing the training signals for planning.}  
Details about training data construction is in \S~\ref{sec:plan_data}. 

At each decoding step, the individual language models, $p(y_t|y_{<t}, \bm{x}_i)$, and 
the distribution scores, $d(\bm{x}_i|y_{<t})$, are first calculated in parallel. We then decode each token greedily based on the mixed language models in an autoregressive way.

\section{Experiment Setups}
\label{sec:exp}
We experiment with the tasks of argument generation and opinion article writing (\S~\ref{sec:data}). Both tasks require generating multi-sentence output, and contain a substantial amount of opinions and factual content. 
We describe the construction of initial content items and the training data for generating expanded concepts and claims in \S~\ref{sec:plan_data}. We present models for comparison in \S~\ref{sec:baseline}. Finally, we provide implementation details in \S~\ref{sec:impl}.  

\subsection{Tasks and Datasets}
\label{sec:data}

\paragraph{Argument Generation.} 
We collect arguments from Reddit ChangeMyView\footnote{\url{https://www.reddit.com/r/changemyview/}} (CMV) community, an online forum that features argumentative discussions. 
Each thread begins with an original post (OP) stating an opinion towards a controversial topic, e.g., ``\textit{The U.S. is too big for one government}''. 
High-quality replies that counter-argue with the OP and are labeled with community endorsement are collected in our prior work~\cite{hua-wang-2020-pair}, covering content posted from 2013 to 2018. In this work, we extend the data collection to 2019. 
Our goal is to generate the entire reply (i.e., the target) given the OP title. 
Statistics about the CMV dataset are listed in Table~\ref{tab:data-stats}. We reserve the most recent $1,000$ samples for test and another $1,000$ for validation. 

\smallskip
\noindent\textbf{Opinion Article Writing.} 
Our second task is to generate opinion articles, as collected from the New York Times (NYT) corpus~\cite{sandhaus2008new}. 
We retain articles whose \texttt{taxonomy} labels include \textit{Top/Opinion}. To ensure that articles can be processed by our computing resource, we only keep the ones with at most 20 sentences, representing $60\%$ of all opinion articles.
As shown in Table~\ref{tab:data-stats}, NYT outputs tend to be significantly longer and contain less claims than CMV. Similarly, we keep $1,000$ examples each for test and validation sets. 

\begin{table}[t]
\small
\centering
\setlength{\tabcolsep}{1.0pt}
\begin{tabular}{l ll}
\toprule
    & {\bf CMV} & {\bf NYT} \\
    \midrule
\# Samples & $77,245$ & $113,616$ \\
Avg. Title Len. & $19.2$ & $5.9$ \\
Avg. \# Cont. Items (\% w/ Claims) & $6.8$ ($76.5\%$) & $9.3$ ($38.9\%$) \\
Avg. \# Core Concepts & $3.6$ & $4.8$ \\
Avg. \# Predicted Concepts & $4.2$ & $4.3$ \\
Avg. \# Entities & $0.8$ & $0.7$ \\
Avg. Target Generation Len. & $142.0$ & $218.9$ \\
Cov. by Core Concepts & $13.2\%$ & $14.9\%$ \\
Cov. by Augmented Concepts & $16.9\%$ & $18.7\%$ \\
Cov. by Augmented Cont. Items & $52.4\%$ & $39.1\%$ \\
\bottomrule
\end{tabular}
    \caption{
  Statistics of the two datasets. We report average numbers of concepts and entities per content item, and the coverage of words in target generations by different input options.
  }
  \label{tab:data-stats}
\end{table}

\subsection{Content Item Construction}
\label{sec:plan_data}

From target references, we describe how to automatically construct the input content items consisting of entities and core concepts, and how to collect training data to fine-tune BART to predict more specific concepts and additional claims. 
Prior work has demonstrated the benefits of incorporating knowledge bases for text generation~\cite{clark-etal-2018-neural,puduppully-etal-2019-data,guan-etal-2020-knowledge}.  
We thus consider two sources of knowledge: (1) entities from Wikipedia, which are useful for modeling events and opinion targets, and (2) concept words from ConceptNet~\cite{speer2017conceptnet}, that cover more related details. 
Note that our setup is generally applicable to other text generation tasks, as these input items can be obtained through standard NLP pipelines, as described below.

\smallskip
\noindent\textbf{Entity Linking.} 
We first segment a reference into sentences. The ones with fewer than $5$ tokens are discarded for content item construction. 
For the rest, we extract entity mentions using Stanford CoreNLP~\cite{manning2014stanford}, and further include nominal noun phrases. 
For entity linking, we adopt CrossWiki~\cite{spitkovsky2012lrec}, which can process our large-scale data within a reasonable amount of time. CrossWiki maps a mention to a list of frequently linked Wikipedia entries. 
We further manually verify and correct the linking results for the top $500$ most frequent mentions. 

\smallskip
\noindent\textbf{Concept Extraction.}
To identify concepts in a reference, we match the lemmatized unigrams and their part-of-speech (POS) tags against all ConceptNet entries.
To create a reasonably challenging task, we only keep a subset of the matches for inclusion in the core concept set (i.e., $C_i$), with the rest used as $C^+_i$, to be generated by our concept expansion model.
Furthermore, we conjecture that an opinion article author tends to start with high-level topics that cover more abstract topical words.
We thus leverage a lexicon~\cite{brysbaert2014concreteness} with concreteness scores, ranging from 0 (abstract) to 5 (concrete), for over $40$k English words. 
We keep concepts that are verbs or have a concreteness score lower than $3.0$.
Word coverage of references by using core concepts and additionally with augmented concepts are $13.2\%$ and $16.9\%$ on CMV respectively, and similarly on NYT (Table~\ref{tab:data-stats}). Finally, we \textbf{train a concept generator} with BART to produce $C^+_i$, conditional on $C_i$, the title, and the entities.

\smallskip
\noindent\textbf{Claim Detection and Generation.} 
Claims are indispensable for opinion articles. As described in \S~\ref{sec:encoding}, we aim to enrich content items with claims targeting the given entities within the title's context. 
To this end, we first \textbf{train a claim detector} by fine-tuning a BERT$_\text{base}$~\cite{devlin-etal-2019-bert} sequence classifier with a dataset consisting of sentences of \texttt{claims} and \texttt{facts}. 
Concretely, we collect $54,802$ claim sentences from Kialo\footnote{\url{https://www.kialo.com/}}, a repository for debate arguments. We then sample $50,000$ sentences from Wikipedia, which are treated as facts. 
This classifier is applied on a reference, and sentences that are labeled as claims become the target for our claim generator. 

We then \textbf{learn a claim generator} using BART, which takes in the title and the entities, and outputs the claim. We augment our training data with replies collected from $30$ active subreddits related to political discussions, with details in Appendix~\ref{sec:appendix-claim}. In total, $80,566$ sentences, which contain at least one entity and are labeled by our classifier as \texttt{claim}s, are kept to train the generator.






\begin{table*}[t]
\centering
\fontsize{10}{12}\selectfont
\small
 \setlength{\tabcolsep}{1.5mm}
\centering
    
    \begin{tabular}{l D{,}{.}{2.2}D{,}{.}{2.2}D{,}{.}{2.2}r l D{,}{.}{2.2}D{,}{.}{2.2}D{,}{.}{2.2}r}
    \toprule
        & \multicolumn{4}{c}{Argument Generation (CMV)} & \phantom{} & \multicolumn{4}{c}{Opinion Article Generation (NYT)} \\
        
        & \multicolumn{1}{c}{\bf{BLEU-2}} & \multicolumn{1}{c}{\bf{ROUGE-2}} & \multicolumn{1}{c}{\bf{METEOR}} & \multicolumn{1}{c}{\bf{Len.}} & \phantom{} & \multicolumn{1}{c}{\bf{BLEU-2}} & \multicolumn{1}{c}{\bf{ROUGE-2}} & \multicolumn{1}{c}{\bf{METEOR}} & \multicolumn{1}{c}{\bf{Len.}} \\
        \midrule
        \textsc{Retrieval} & 6,29 & 3,68 & 10,00 & 78 & \phantom{} & 9,68 & 7,96 & 9,98 & 99 \\ 
        \textsc{SentPlanner} & 7,78 & 3,23 & 7,69 & 114 & \phantom{} & 7,45 & 5,06  & 6,62  & 106 \\ 
        \textsc{Seq2seq} & 16,71 & 9,53 & 13,34 & 100 & \phantom{} & 21,44 & 14,92 & 14,93 & 119 \\
        \textsc{Seq2seqFull} & 29,11 & 17,71 & 20,27 & 145 & \phantom{} & 31,06 & 29,74 & 23,10 & 121 \\ 
        \midrule
        \textsc{DYPLOC} (ours)  & {\bf 32},{\bf 60} & {\bf 25},{\bf 69} & {\bf 22},{\bf 61} & 101 & \phantom{} & {\bf 40},{\bf 63} & {\bf 36},{\bf 93} & 25,76 & 122  \\
        \quad w/o All Concepts & 7,80 & 3,68 & 7,21 & 107 & \phantom{} & 11,32 & 6,01 & 8,33 & 132 \\
        \quad w/o Augmented Concepts & 22,39 & 15,90 & 16,91 & 99 & \phantom{} & 26,94 & 21,56 & 18,39 & 117 \\
        \quad w/o Claims & 31,62 & 25,03 & 22,09 & 100 & \phantom{} & 39,44 & 35,43 & 25,25 & 122 \\ 
        \quad w/o Entities & 32,11 & 25,36 & 22,42 & 101 & \phantom{} & 39,66 & 35,82 & 25,11 & 122 \\ 
        \hdashline
        \quad Random Selection & 12,96 & 8,25 & 10,05 & 103 & \phantom{} & 5,32 & 5,29 & 6,00 & 72 \\ 
        \quad Greedy Selection & 32,33 & 25,60 & 22,53 & 100 & \phantom{} & 40,61 & 36,88 & {\bf 25},{\bf 77} & 122 \\
        \bottomrule
    \end{tabular}
    \caption{
    Automatic evaluation results on both tasks. We report BLEU-2, ROUGE-2, METEOR, and output length. Best scores are in bold. Our \textsc{DYPLOC} model statistically significantly outperforms all baselines and comparisons (randomization approximation test~\cite{noreen1989computer}, $p < 0.0005$).
  }
    \label{tab:main-results}
\end{table*}

\subsection{Baselines and Comparisons}
\label{sec:baseline}

We compare with three baselines: (1) \textsc{\textbf{Retrieval}} first calculates the TF-IDF weighted bag-of-words vectors for each content item, which is then used to query the training set sentences. 
The one with the highest cosine similarity is picked for each query, which are then ordered by a trained Pointer-Network~\cite{NIPS2015_29921001} as described in \newcite{gong2016end}. 
(2) \textsc{\textbf{SentPlanner}}~\cite{hua-wang-2019-sentence} is an LSTM-based seq2seq model with a separate sentence planning decoder, where the planner selects keyphrases by using attentions and the generator reflects the selections. We treat our entities and concepts as keyphrases to feed to this model. 
(3) \textsc{\textbf{Seq2seq}} is a fine-tuned BART model, whose input is the original content items \textit{without augmentation}, thus does not have access to the predicted concepts and claims. 

Additionally, we consider a strong comparison \textsc{\textbf{Seq2seqFull}}, by fine-tuning BART with the same augmented content items as inputs as in our model. The difference is that the content items are concatenated before being used as input.

\subsection{Reproducibility}
\label{sec:impl}

We implement all models using the Huggingface Transformers library~\cite{wolf-etal-2020-transformers} with PyTorch~\cite{paszke2019pytorch}.
We use the base model for BART, which has $768$ dimensional states and $6$ layers for both encoder and decoder ($140$M parameters in total). Our newly added plan scoring network only contains $1.2$M parameters, less than $1\%$ of the pre-trained model.
Our generation model is optimized using Adam~\cite{kingma2014adam}, with a batch size of $3$. 
To improve efficiency, we adopt the mixed-precision (FP16) to train each model, using one NVIDIA Titan RTX GPU card with 24GB memory.
The number of content items is limited to $10$ per sample, and the numbers of entities and concepts per content item are capped at $20$, respectively. We also truncate the target output to at most $200$ tokens during training. 
Early stopping is applied over validation loss.
Our model converges after being trained for $38$ hours ($19$ epochs) on CMV, and $45$ hours ($15$ epochs) on NYT. 
The best validation perplexity reaches about $6.1$ after model convergence on both datasets.

\section{Results}
\label{sec:results}
\subsection{Automatic Evaluation}
\label{sec:auto-eval}
Here we report results on test sets with standard automatic metrics: BLEU~\cite{papineni-etal-2002-bleu} measures the n-gram precision (here we consider up to bigrams); ROUGE~\cite{W04-1013}, calculated based on n-gram recall; and METEOR~\cite{denkowski-lavie:2014:W14-33}, which also accounts for synonyms. 
In Table~\ref{tab:main-results}, we first present the results when gold-standard concept expansion is used. 

Our proposed \textsc{DYPLOC} model achieves significantly higher performance across all metrics on both datasets. 
In particular, the substantial lead over \textsc{Seq2seqFull}, which has access to the same content items as ours, indicates that \textit{dynamic content planning with mixed language models produces superior generations}. 
Among comparison models, the gap between \textsc{Seq2seqFull} and \textsc{Seq2seq} shows the effectiveness of content item augmentation. We also observe a significant drop for baselines without using large models, highlighting the importance of pre-training. 

\begin{table}[t]
    \centering
    \fontsize{10}{11}\selectfont
    \small
 \setlength{\tabcolsep}{0.5mm}
    \begin{tabular}{l D{,}{.}{2.2}D{,}{.}{2.2} D{,}{.}{2.2}D{,}{.}{2.2}}
        \toprule
        & \multicolumn{2}{c}{CMV} & \multicolumn{2}{c}{NYT} \\
         &  \multicolumn{1}{c}{{\bf BLEU-2}} & \multicolumn{1}{c}{{\bf METEOR}} & \multicolumn{1}{c}{{\bf BLEU-2}} & \multicolumn{1}{c}{{\bf METEOR}} \\
         \midrule
        \textsc{Retrieval} & 8,30 & 9,64 & 8,85 & 9,21 \\
        \textsc{SentPlanner} & 7,84 & 7,76 & 7,75 & 6,80 \\
        \textsc{Seq2seqFull} & 18,06 & 15,96 & 16,20 & 15,25 \\
        \textsc{DYPLOC} & {\bf 22},{\bf 84} & {\bf 17},{\bf 13} & {\bf 24},{\bf 54} & {\bf 17},{\bf 41} \\
        \bottomrule
    \end{tabular}
    \caption{BLEU-2 and METEOR (MTR) results on systems with predicted concepts as input. Same trends are observed on ROUGE, which are in Appendix~\ref{sec:appendix-eval}.}
    \label{tab:system}
\end{table}

\smallskip
\noindent\textbf{Ablation Study.} 
To verify the effect of each element in content items, we further train ablated models by removing concepts, claims, or entities.
The results are also displayed in Table~\ref{tab:main-results}. 
In general, scores decrease when using only partial content items, among which removing all concepts lead to the biggest performance drop, suggesting that entities and claims alone are insufficient to produce informative outputs.

\smallskip
\noindent\textbf{Effect of Hard Selection of Content Items.} 
To test the necessity of using weighted-sum marginalization (Eq.~\ref{eq:marginalization}), we experiment with two comparisons with hard selections, i.e., either randomly choosing a content item, or using the one with the highest predicted plan score (greedy selection). 
For both cases, we set the selected content item's plan score as $1.0$, with the rest of the candidates having a score of $0.0$, to ensure the probabilities summed up to $1.0$. 
As can be seen from the bottom two rows of Table~\ref{tab:main-results}, not surprisingly, random selection performs much worse. We observe that its generations lack coherence and fluency, implying the effectiveness of our learnable content planner. 
On the other hand, using greedily selected content items obtains comparable results with \textsc{\method}, where a weighted sum of content items is considered. Indeed, we find that \textsc{\method}'s plan scores are often sharp where one content item has much higher weight than others, and in these scenarios, it is almost equivalent to the greedy selection setup. 

\smallskip
\noindent\textbf{Results with Generated Concepts.}
Table~\ref{tab:system} lists generation results with our \textit{system generated concepts} as expansion. While all systems yield worse results compared to using gold-standard concepts, our \textsc{DYPLOC} still outperforms other models by substantial margins, showing its \textit{robustness when input concepts are noisy}. 
Yet it also suggests the importance of having more accurate and comprehensive concept expansion, which should be explored in the future work.

\definecolor{Gray}{gray}{.8}
\begin{table}[t]
\fontsize{10}{11}\selectfont
\small
 \setlength{\tabcolsep}{.9mm}
  \centering
  
   \begin{tabular}{l l  l l l l l}
    \toprule
    {\bf Data} & {\bf System} & {\bf Gram.} & {\bf Coh.} & {\bf Rel.} & {\bf Cont.} & {\bf Top-1} \\
    \midrule
    
      CMV & \textsc{Seq2seq} & $4.19$ & $3.12$ & $3.19$ & $2.89$ & $25.1\%$ \\
          & \textsc{Seq2seqFull} & $4.24$ & $3.19$ & $3.23$ & $3.13$ &  $30.2\%$ \\
          & \textsc{\method} & ${\bf 4.26}$ & ${\bf 3.35}$ & ${\bf 3.35}$ & ${\bf 3.28}$ & ${\bf 44.7\%}$\\
    \midrule
 
     NYT & \textsc{Seq2seq} & $4.38$ & $3.82$ & $4.20$ & $4.01$ & $25.2\%$ \\
         & \textsc{Seq2seqFull} & $4.48$ & $3.99$ & $4.30$ & $4.14$ & $28.9\%$\\
         & \textsc{\method} & ${\bf 4.55}$ & ${\bf 4.14}$ & ${\bf 4.31}$ & ${\bf 4.28}$ & ${\bf 45.9\%}$ \\
    \bottomrule
      
   \end{tabular}
  \caption{
  Human evaluation results on grammaticality (Gram.), relevance (Rel.), coherence (Coh.), and content richness (Cont.). For each sample, outputs by all three systems are ranked based on the overall preference. We show the percentage each system is ranked as the best. 
  }
  \label{tab:human-eval}
 \end{table}
 
 \begin{table*}[t]
\centering
\fontsize{9}{11}\selectfont
        \setlength{\tabcolsep}{.1mm}
        \begin{tabular}{p{20mm} p{140mm}}
        \toprule
        
        \textit{Content Items} & \hlyellow{(a)} \texttt{[ENT]} CO2 \texttt{[CON]} death, ensue, \textit{toll}, \textit{staggering} \hlred{(b)}\texttt{[CON]} leave, stop, compare, change, denialism, issue, \textit{simply}, \textit{risk} \hlblue{(c)}\texttt{[ENT]} Fossil\_fuel \texttt{[CON]} drive, paralyze, deal, \textit{humanity}, \textit{industry} \texttt{[CLAIM]} \textit{Coal is not a reliable source of energy, and it's been driven by unreliable sources of energy and unreliable sources to date.} \hlgreen{(d)}\texttt{[CON]} win, consequence, remotely, \textit{dire}
 \\
        
        \textsc{DYPLOC} & \hlblue{Climate change denial is driven by fossil fuel industries that are paralyzing the humanities in attempting to deal with the problem.} \hlyellow{The death toll from ensuing CO2 is staggering.} \hlgreen{There is no winning consequence for anything remotely dire that climate change denialism has.} \hlred{Leaving aside the issue of GM risk, simply comparing climate change denialism to climate change protesters is not going to stop anything.}  \\
        
        \textsc{Seq2seqFull} & World wide civilisation is threatened with catastrophe by climate change denial. Climate change denial is driven by the fossil fuel industry, which is paralyseed by the humanities. Any attempt to deal with the problem of climate change is going to cause death tolls that ensue, staggering. Leaving aside the issue of GM, the risk is is simply too much to compare. \\
        
        \midrule
        \end{tabular}
      
      \begin{tabular}{p{20mm} p{140mm}}

        \textit{Content Items} & \hlyellow{(a)}\texttt{[ENT]} President\_of\_the\_U.S., George\_W.\_Bush, Richard\_Nixon \texttt{[CON]} omit, \textit{list} \hlred{(b)}\texttt{[ENT]} George\_W.\_Bush, \texttt{[CON]} bring, people, world, \textit{honor}, \textit{dignity}, \textit{respect} \hlblue{(c)} \texttt{[ENT]} Bill\_Clinton, Ronald\_Reagan \texttt{[CON]} harm, respect, demonstrate, damage, \textit{office}, \textit{state} \hlgreen{(d)} \texttt{[CON]} owe, correction, omission, \textit{apology}, \textit{readers} \\
        
        \textsc{DYPLOC} & \hlblue{President Clinton ``has done more harm to the respect of the Presidency than Presidents Presidents Ford, Carter and Reagan have demonstrated,'' and that ``the failure of Presidents Clinton and Reagan has caused permanent damage to the office.''} \hlyellow{You omit President Bush from your list of Presidents.} \hlred{Mr. Bush has brought honor, dignity and respect to the office of the Presidency in the eyes of the American people and the world at large.} \hlgreen{We owe your readers an apology and a correction of your editorial's glaring omission, which created an erroneous impression.}  \\
        
        \textsc{Seq2seqFull} & ``The Picture of Bill Clinton'' (editorial, Dec. 5) states that President Clinton ``has done more harm to our respect for the Presidency than Presidents Reagan, Ford, Carter and Bush.'' This demonstrates that President Reagan's failure to do more damage to our honor than President Bush's failure in office. You omitted from your list President Clinton's achievements that brought honor and dignity to the eyes of the American people and to the world at large.  [...] \\
        
        \bottomrule
        \end{tabular}
\caption{\fontsize{10}{12}\selectfont 
Sample generations on CMV [Upper] and NYT [Lower]. System generated concepts and claims are in \textit{italics}.
For DYPLOC, we highlight sentence to content item alignment using colors.
}
\label{tab:sample-output}
\end{table*}

\subsection{Human Evaluation}
\label{sec:human}

We hire three proficient English speakers to evaluate four key aspects of the generated outputs: 
(1) \textbf{grammaticality}; (2) \textbf{coherence}, measuring if the text is logical and cohesive; (3) \textbf{relevance}, gauging topic relatedness to the input title; and (4) \textbf{content richness}, assessing the specificity and whether there is enough details in the outputs. 
Each aspect is rated on a scale of 1 (worst) to 5 (best). In addition, judges also rank the system outputs by their overall preferences. 
Detailed evaluation guideline is attached in Appendix~\ref{sec:appendix-human}. 

We randomly select $50$ samples from the test sets for both tasks, and present outputs by \textsc{Seq2seq}, \textsc{Seq2seqFull}, and \textsc{DYPLOC} in random orders. 
Table~\ref{tab:human-eval} shows that \textsc{DYPLOC} receives higher scores across all aspects and tasks. In particular, the considerable differences in \textbf{coherence} and \textbf{content richness} indicate that \textit{our framework yields better content organization as well as retains more useful information}. Overall, our system outputs are ranked best for $44.7\%$ and $45.9\%$ of the time in two tasks, significantly more than the comparisons.

\smallskip
\noindent\textbf{Analysis on Argumentative Quality.}
In the ablation study, we find that our full model's performance is similar to the version without having claims as input. We suspect this is because claims are often paraphrased or even not directly used when delivering an argument, which cannot be captured by the automatic metrics. 
To better understand how claims are used for generation, we randomly select $50$ examples by \textsc{DYPLOC} and its variant without claims, and ask the same human judges to decide whether there is a clear central argument conveyed by each generated argument on CMV. 

We observe that $66.7\%$ of the outputs by our full model are recognized as \textit{successfully delivering arguments with consistent stances}, whereas only $61.3\%$ are true for the model variant without claims. 
This gap confirms that claim drafts can indeed promote the argumentative quality as perceived by human readers.

\section{Further Discussions}
\label{sec:discussion}

Evaluation results on generation quality have shown the effectiveness of our mixed language models. In this section, we aim to further understand the behavior of the plan scoring network, $d(\bm{x}|y_{<t})$, such as how it affects the usage of content items for generation. 
Specifically, we adopt the following procedure to construct \textbf{alignment} between each sentence in the output and content items: for each token $y_t$, 
we establish a mapping $y_t \mapsto \bm{x}_i$ if $\bm{x}_i$ is the most important item for producing $y_t$, i.e., $\bm{x}_i = \argmax_{\bm{x}} d(\bm{x}|y_{<t})$, and $d(\bm{x}_i|y_{<t}) > 0.5$. If all tokens in an entire sentence are mapped to the same $\bm{x}_i$, we consider this sentence is aligned to that content item. Based on this rule, we show sample output and corresponding alignments in Table~\ref{tab:sample-output}.

For the rest of this section, we conduct analyses based on this alignment result. 
We first examine whether the model learns to utilize enough content items, i.e., high coverage. Then we provide insights on whether the generation faithfully reflects the argumentative claims using entailment relation labeling by human inspection.

\begin{figure}[t]
    \pgfplotsset{height=4cm,width=7cm,compat=1.6,
                /pgfplots/ybar legend/.style={
                /pgfplots/legend image code/.code={%
                \draw[##1,/tikz/.cd,yshift=-0.25em]
                (0cm,0cm) rectangle (3pt,0.8em);},
    },}
\centering   
\begin{tikzpicture}
    \begin{axis}[
    title={Content Item Coverage by Output (\%)},
    title style={at={(0.5, 0.9)}, font=\small},
    xticklabel style={at={(0.5, -0.5)}, font=\small},
    ybar=10pt,
    ymin=0,
    ymax=120,
    ytick={0, 25, 50, 75, 100},
    enlarge x limits=0.4,
    enlarge y limits=0,
    legend style={at={(0.5,-.25)},
    anchor=north,legend columns=-1, font=\small},
    symbolic x coords={CMV, NYT},
    xtick=data,
    bar width=12pt,
    nodes near coords,
    every node near coord/.append style={font=\small},
    nodes near coords align={vertical},
    xtick pos=left,
    ytick pos=left,
    grid=major,
    ]
        \addplot coordinates {(CMV,87.25) (NYT,83.89)};
        \addplot coordinates {(CMV,96.71) (NYT,90.87)};
        \legend{{w/ Generated Concepts}, {w/ Oracle Concepts}}
    \end{axis}
\end{tikzpicture}

\caption{The percentage of content items that are aligned to at least one output sentence.}
\label{fig:content_coverage}
\end{figure}
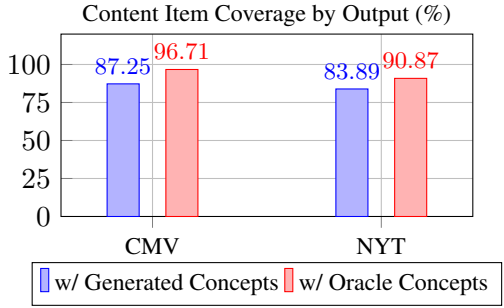

\smallskip
\noindent\textbf{How many content items are used by the output?} 
Human judges have rated our model output to contain more relevant information (Table~\ref{tab:human-eval}). We believe this can be attributed to the enhanced capacity to access and reflect the input data with dynamic content planning, as a result of mixed language models. To verify this hypothesis, we calculate the percentage of content items that are aligned to at least one output sentence. 
Figure~\ref{fig:content_coverage} shows that, using our system, the coverage reaches $87.25\%$ on CMV and $83.89\%$ for NYT. If we replace the generated concepts with gold-standard concepts (as extracted from references) instead, the coverage exceeds $90\%$ on both tasks. 
These observations indicate that \textit{our model can indeed adequately utilize the input data, with more accurate concepts further encouraging higher coverage}. 

\smallskip
\noindent\textbf{How are claim content items realized?}
Claims are the central elements for opinion text construction. 
As mentioned in \S~\ref{sec:plan_data}, a subset of the content items are supplied with claim sentences.
In order to examine whether they are realized as claim sentences in the outputs, we leverage the fine-tuned BERT classifier (\S~\ref{sec:plan_data}) to label all output sentences. $90.96\%$ of the sentences that are aligned to a claim element in the input are also labeled as \texttt{claim} on CMV. The percentage is only $69.41\%$ for NYT, though, likely because the NYT opinion articles still contain more objective information. 

Furthermore, we conduct a human evaluation study to assess the semantic relations between claim input and its aligned generated sentence.
We randomly sample $50$ outputs from test sets, and ask four human judges to read each. 
For each sample, we highlight one output sentence that is aligned to a content item with claim element. The judges determine a three-way (\textsc{entail}, \textsc{neutral}, \textsc{contradictory}) entailment relation between the input claim (premise) and the output (hypothesis). Results show that \textsc{entail} accounts for $49.06\%$ of all instances, while only $3.77\%$ are deemed \textsc{contradictory}. Upon inspection, the contradictory pairs are usually disagreements with regard to implicit sentiments, e.g., ``\textit{Journalist is the most responsible for the problem}'' vs. ``\textit{Media coverage is a good thing.}''.  
This suggests that while our conditional language model achieves reasonable semantic control in most cases, it is still not guaranteed to capture more nuanced semantics encoded in opinions and arguments.  
Future work includes designing representations that can better model stances in opinions as well as argumentative structures. 

\section{Conclusion}
\label{sec:conclusion}
We present a novel text generation framework that enables dynamic content planning based on mixed conditional language models. We further employ large models to augment system inputs with diverse content that covers both objective and subjective information. 
The experiments on two distinct opinion text generation tasks show that our proposed model compares favorably against strong comparisons based on fine-tuned BART models with the same input. Human evaluation further confirms that our model generations have richer information and better content organization. 

\section*{Acknowledgements}
This research is supported in part by National Science Foundation through Grant IIS-1813341. 
We thank three anonymous reviewers for their valuable suggestions on various aspects of this work. 

\section*{Ethics Statement}
\label{sec:ethics}
Large models that are pre-trained on heterogeneous web data are shown to encode biases and can be potentially harmful for marginalized populations. 
Along with the improved controllability, we also recognize that our system might be misused to create fabricated or offensive content. We therefore advocate cautious and responsible practices in real-world deployment.

\bibliographystyle{acl_natbib}
\bibliography{reference}

\appendix
\section{Training Data Construction for Claim Generator}
\label{sec:appendix-claim}
We describe the claim generation model in \S~\ref{sec:plan_data} for content item enrichment. Since both our CMV and NYT data focus on the politics domain, we leverage a collection of Reddit posts from politics related subreddits. The full list of subreddits are shown in Table~\ref{tab:subreddits}. In total, we collect $1.6$ million posts, which are split into sentences, among which we only keep the ones classified as \texttt{claim} by the BERT$_\text{base}$ classifier and have at least one named entity.

\begin{table}[h]
\centering
\fontsize{9}{11}\selectfont
\setlength{\tabcolsep}{1mm}
\begin{tabular}{|ll|}
    \hline
     \texttt{Anarchism} & \texttt{AmericanPolitics}  \\
     \texttt{Capitalism}  & \texttt{Anarcho\_Capitalism} \\
     \texttt{Conservative} & \texttt{democracy}\\
     \texttt{democrats} & \texttt{feminisms} \\
     \texttt{government} & \texttt{GreenParty} \\
     \texttt{IWW} & \texttt{labor} \\
     \texttt{Liberal} & \texttt{Libertarian} \\
     \texttt{LibertarianLeft} & \texttt{LibertarianSocialism} \\
     \texttt{Marxism}  & \texttt{moderatepolitics} \\
     \texttt{Objectivism}  & \texttt{PoliticalDiscussion} \\
     \texttt{politics}  & \texttt{progressive} \\
     \texttt{Republican}  & \texttt{republicans} \\
     \texttt{socialdemocracy} & \texttt{socialism} \\
     \texttt{ukpolitics} & \texttt{uspolitics} \\
     \texttt{worldpolitics} & \texttt{PoliticalPhilosophy} \\
     \hline
\end{tabular}
\caption{List of subreddits used to construct training data for learning the claim generator.}
\label{tab:subreddits}
\end{table}

\section{Additional Automatic Evaluation Results}
\label{sec:appendix-eval}

In \S~\ref{sec:auto-eval}, we report results by automatic metrics using system predicted concepts in Table~\ref{tab:system}. Here we additionally show the results evaluated by ROUGE-2 and average output lengths in Table~\ref{tab:system-2}.

\begin{table}[t]
    \centering
    \fontsize{10}{12}\selectfont
    \small
 \setlength{\tabcolsep}{1.4mm}
    \begin{tabular}{l D{,}{.}{2.2}r D{,}{.}{2.2}r}
        \toprule
        & \multicolumn{2}{c}{CMV} & \multicolumn{2}{c}{NYT} \\
         &  \multicolumn{1}{c}{{\bf ROUGE-2}} & {\bf Len.} & \multicolumn{1}{c}{{\bf ROUGE-2}} & {\bf Len.} \\
         \midrule
        \textsc{Retrieval} & 4,39 & 82 & 6,64 & 95 \\
        \textsc{SentPlanner} & 3,24 & 115 & 5,12 & 108 \\
        \textsc{Seq2seqFull} & 8,83 & 120 & 8,83 & 135 \\
        \textsc{DYPLOC} & {\bf 11},{\bf 83} & 118 & {\bf 15},{\bf 46} & 134 \\
        \bottomrule
    \end{tabular}
    \caption{ROUGE-2 and average length (Len.) on systems with predicted concepts as input.}
    \label{tab:system-2}
\end{table}


\section{Human Evaluation Guideline}
\label{sec:appendix-human}
We include the detailed human evaluation guidelines in Figure~\ref{tab:human-eval-1}. Note that we collect $53$ samples for annotation for each domain. The first three are for calibration only and not be included in the final results. 

\begin{figure*}[h]
\centering
\scalebox{0.95}{
\begin{tikzpicture}

    \node[draw, inner sep=10pt, rounded corners]{
    \centering\fontsize{9}{11}\selectfont
    \begin{tabular}{p{135mm}}

         In the following studies, you will evaluate the system outputs of three text generation models on two different domains. For each domain, there will be $53$ examples presented, each starting with a statement, followed by three system generations. 
         Please first read the statement and then the system outputs. At the end of each output, please provide your judgment on the quality of the following aspects, based on a scale of 1 (worst) to 5 (best):

        {\fontsize{9}{11}\selectfont
        \begin{itemize}[leftmargin=3mm]
            \item {\bf Grammaticality:} whether the text reads fluently and has no grammar error
            {\fontsize{9}{11}\selectfont
            \begin{itemize}[leftmargin=3mm]
                \item 1. Major grammatical errors that significantly impact comprehension of text. E.g., \textit{``I'm not a quick skimming, but im quickly making a comment.''}.
                
                \item 3. Minor grammatical errors that do not significantly impact comprehension of text. E.g., \textit{``I have car that works, and I make it to work by commute 45 minutes to an hour on my bike.''}
                
                \item 5. No grammatical issues. E.g., \textit{``There are swathes of people whose function is determined by technology, and they use technology as a crutch.''}
                
            \end{itemize}}
            \item {\bf Coherence:}  whether the information transition is natural and well-structured
            \begin{itemize}[leftmargin=3mm]
                \item 1.  Sentences are completely unrelated. E.g., \textit{``The Supreme Court created a mechanism for interpreting the Constitution through a modern lens. The question is, do you create jobs? Ukraine is a direct ally of the US.''}

                \item 3. Sentences are connected to one another but transitions seem disjointed; there doesn't appear to be a strong progression of ideas. E.g., \textit{``Muslims worship the figure of Allah. Christians worship the figures of God. Muslims do not worship the Jews. Muslims don't worship the Christian figure of God, Muslims worship God. They worship the Jewish figure of the figure.''}

                \item 5. Sentences transition smoothly and connect to form a logical progression. E.g., \textit{``Every country has to deal with their own geography. USA benefits from decent climate country wide, plentiful natural resources and distance from areas of war. The downside is that they are close to Mexico and Mexico pretty much sucks, so it's inhabitants want to get into the USA. Unless you believe that all resources and other benefits should be shared then why should the world take on the USA downfalls while not getting any of the plusses?''}

            \end{itemize}
            \item {\bf Relevance:} whether the content of the text is relevant to the title 
            \newline \quad\quad Title: \textit{The recent swell in protesting Commencement speakers at colleges is a good thing.}
            \begin{itemize}[leftmargin=3mm]
                \item 1. The output is generic or completely irrelevant. E.g., \textit{``Supply and demand. The US thinks those drugs are worth price X. Other countries are only willing to pay price Y. The US develops more IP related content than other countries because it has a huge military and is able to enforce IP laws.''}

                \item 3. The text is tangential to the title and the input (it may share an entity or key concept in common), though it might not be precisely on topic. E.g., \textit{``When you enter a college career, you decide to take literature studies. You can become an engineer, history, linguistics, etc.''}

                \item 5. The text is highly relevant with the title and the input. E.g., \textit{``The problem with protesting minority opinions is that you force the majority opinion to come out against them, and as a result you find controversial speakers turning their commencement speeches into bland speeches. Commencement speeches are a recognition of a person, and offer an affirmation of their worldview.''}
                \end{itemize}
            \item {\bf Content Richness:} whether the output conveys substantial content
            \begin{itemize}[leftmargin=3mm]
                \item 1. Generic response with no useful information about the topic. E.g., \textit{``I don’t have time to address the point you're making.''}

                \item 3. With one or two key points that are useful as counter-argument. E.g., \textit{``Reducing costs is not the goal of the free market.''}

                \item 5. With sufficient key information that is useful as counter-argument. E.g., \textit{``Reducing costs is not the goal of the free market. Simply setting prices for medical procedures has been shown to be extremely effective. I will tacitly admit that your post is true for many countries, but the US health share is less than 1\% of GDP.''}
                \end{itemize}
        \end{itemize}} 
         \\
    
    \end{tabular}};

\end{tikzpicture}}
    \caption{Human evaluation guidelines and representative examples on rating scales.}
    \label{tab:human-eval-1}
\end{figure*}

\section{Additional Sample Outputs}

Additional example content items and generations are demonstrated in Table~\ref{tab:sample-output-cmv} and Table~\ref{tab:sample-output-nyt}.

\begin{table*}[t]
\centering
\fontsize{9}{11}\selectfont
        \setlength{\tabcolsep}{.1mm}
        \begin{tabular}{p{20mm} p{140mm}}
        \toprule
        
        \textit{Title} &  I feel that humanity is going backwards due to a rise in internet use and lack of face to face communication.  \\
        \textit{Link} & \url{https://www.reddit.com/r/changemyview/comments/1xklhm} \\
        \midrule
        \textit{Content Items} & \hlyellow{(a)} \texttt{[CON]}  give, coherent, impossible, argue, \textit{argument}, \textit{view} \hlred{(b)}\texttt{[CON]} intent, normal, face, conversation, interpret, body, language, \textit{facial}, \textit{gesture}, \textit{pitch}, \textit{tone}, \textit{content} \hlblue{(c)}\texttt{[ENT]} Internet \texttt{[CON]} true, relate, assertion, change, \textit{advent}, \textit{stop}, \textit{move}, \textit{forward} \texttt{[CLAIM]} \textit{I do believe that the internet has been beneficial in the past for communication, but I believe in a society where we are constantly moving forward and we can always adapt} \hlgreen{(d)}  \texttt{[CON]} modern, affect, attitude, communication, generation, current, future, \textit{society}, \textit{skill}, \textit{relationship}
 \\
        
        \textsc{DYPLOC} & \hlgreen{I don't think modern society is affecting attitudes, communication skills, or relationships in current and future generations.}
        \hlred{I think the intent of normal face to face conversations is to interpret body language, facial gestures, pitch tone, content, etc.}
        \hlblue{This is true, but it is not related to your assertion that the changes caused by the advent of the internet stop us from moving forward.}
        \hlyellow{You haven't given a coherent argument for your view, and it is impossible to argue against it.}  \\
        
        \textsc{Seq2seqFull} &  Modern society is affected by the attitudes of communication skills and relationships of current and future generations. 
        The intent of normal face to face conversation is to interpret body language, facial gestures, pitch and tone content. 
        This is true of all communication related to the internet. 
        Your assertion that society is changing because of the advent of the internet is not true. It's not that society has stopped moving forward, it's that society hasn't moved forward. I can't give a coherent argument for your view, because it's impossible to argue against it. \\
        
        \midrule
        \end{tabular}
      
      \begin{tabular}{p{20mm} p{140mm}}
       
        \textit{Title} & I believe all police in the UK should be armed with both tasers and guns. CMV \\
        \textit{Link} & \url{https://www.reddit.com/r/changemyview/comments/1z4s02} \\
        \midrule
        
        \textit{Content Items} & \hlyellow{(a)}\texttt{[ENT]} Police, Alternative\_rock \texttt{[CON]} function, prevent, idea, force, literally, \textit{crime}, \textit{military}, \textit{founding} 
        \hlred{(b)}\texttt{[ENT]} Gun \texttt{[CON]}  comment, argue, agree, important, \textit{aspect}, \textit{consent}, \textit{principles} 
        \hlblue{(c)} \texttt{[ENT]} Police, Filename\_extension \texttt{[CON]} separate, law, completely, power, \textit{elevate}, \textit{people}
        \hlgreen{(d)} \texttt{[CON]} protect, group, people, pay, full, time, \textit{citizen}, \textit{responsibility}, \textit{basically} \\
        
        \textsc{DYPLOC} & \hlyellow{The founding idea of the police force was literally to function as an alternative to preventing crime, and as such, as a military force.} 
        \hlgreen{Basically, a group of people who are paid full time as US citizens have a responsibility to protect themselves.} 
        \hlblue{Separating the police from the law completely elevates the power of the people. }
        \hlred{A lot of the comments argue that guns are necessary, but I agree that the most important aspect of police consent is the principles of consent.} \\
        
        \textsc{Seq2seqFull} & I see a lot of comments arguing for guns, and I agree with you that the most important aspect of police is consent. The principles of the founding of the idea of the police force are literally the only ones that function in preventing crime. The military is basically an extension of the Police force. It's a group of people who pay full time citizens to take responsibility for protecting themselves. Separating them from the law completely elevates their power  \\
        
        \midrule
        \end{tabular}
        
        \begin{tabular}{p{20mm} p{140mm}}
       
        \textit{Title} & I do not believe democracy inherently protects freedom. CMV.  \\
        \textit{Link} & \url{https://www.reddit.com/r/changemyview/comments/1yg8sx} \\
        \midrule
        
        \textit{Content Items} & 
        \hlyellow{(a)}\texttt{[ENT]} Democracy, Democratic\_Party\_(United\_States) \texttt{[CON]} respects, grow, modern, freer, stride, freedom, \textit{century}, \textit{democratic}
        \hlred{(b)} \texttt{[CON]}  modern, democracy, job, protect, freedom, \textit{government}, \textit{people}
        
        \\
        
        \textsc{DYPLOC} & \hlyellow{Modern democracies do a better job of protecting people's freedom from government.} 
       \hlred{In that respect, modern democracies have grown freer over time, and the strides in freedom that took place in the 20th century were mostly because of the democratic movement.} \\
        
        \textsc{Seq2seqFull} & Modern democracy's job is to protect people's freedom. The government in many respects has grown freer over time, and has taken on a different striding place in the 20th century. The democratic movement was formed during the time of the time, during the 2020th century, and during the 21st century, the democratic movement has taken over. The movement has done a better job of protecting freedom than modern democracy.  \\
        
        \bottomrule
        \end{tabular}

\caption{\fontsize{10}{12}\selectfont 
Sample content items and system output on CMV. System generated concepts and claims are in \textit{italics}.
}
\label{tab:sample-output-cmv}

\end{table*}

\begin{table*}[t]
\centering
\fontsize{9}{11}\selectfont
        \setlength{\tabcolsep}{.1mm}
        \begin{tabular}{p{20mm} p{140mm}}
        \toprule
        
        \textit{Title} &  Arafat Visa Refusal  \\
        \midrule
        \textit{Content Items} & \hlyellow{(a)} \texttt{[ENT]} United\_States \texttt{[CON]}  guarantee, review, article, refuse, visa, defy, agreement, explicitly, \textit{editor}, \textit{week}, \textit{access}, \textit{guest}, \textit{nations} 
        \hlred{(b)} \texttt{[ENT]}United\_States, United\_States\_Senate \texttt{[CON]}  give, agreement, limit, construe, ratification, \textit{nations}, \textit{headquarters},  \textit{security}, \textit{reservation}, \textit{safeguard}
        \hlblue{(c)}\texttt{[ENT]} Terrorism, United\_States, Leadership, Palestinian\_people, United\_States\_Department\_of\_State \texttt{[CON]} justify, decision, reject, lend, application, issue, statement, describe, leader, condone,  \textit{support}, \textit{accessory}, \textit{visa} \texttt{[CLAIM]} \textit{The US Department of
        State has determined that terrorism is not a legitimate cause for attack on American soil, and that any attacks on American interests that are not directed at the Israeli people are not justified.} 
 \\
        \textsc{DYPLOC} & \hlyellow{To the Editor: Your May 9 Week in Review article on the United States' refusal to allow a visa to Yasir Arafat defies an agreement explicitly guaranteeing access to United States guests at the United Nations.} 
        \hlblue{To justify the State Department's decision to reject Mr. Arafat's visa application, the United States should issue a statement describing him as a Palestinian leader who condones terrorism, lends support to acts of terrorism and is an accessory to terrorism. }
        \hlred{ratification of the United Nations headquarters agreement would give the United States security reservations that would not be construed as limiting safeguards.}  \\
        
        \textsc{Seq2seqFull} &  To the Editor: Your Dec. 5 Week in Review article on the refusal of a visa to Yasir Arafat defies the agreement explicitly guaranteeing
access to United States guests at United Nations ratification headquarters. The agreement gives security reservations to United Nations guests, but
does not construe them as limiting safeguards to justify the decision to reject the visa application. The State Department issued a statement
describing the decision as a ''leader condoning terrorism.'' The State department has lent support to the Palestinian people by acting as an accessory
to terrorism.  \\
        
        \midrule
        \end{tabular}
      
      \begin{tabular}{p{20mm} p{140mm}}
       
        \textit{Title} &  Can the Global Economy Really Stabilize Itself?  \\
        \midrule
        
        \textit{Content Items} & 
        \hlyellow{(a)}\texttt{[ENT]} Nation \texttt{[CON]} include, impose, limit, foreign, ownership, natural, cumulative, \textit{investor}, \textit{treaty}, \textit{media}, \textit{traditional} \texttt{[CLAIM]} \textit{If the nation is already wealthy and can’t afford to spend its money on its own people, why should it spend their
money on other people?}
        \hlred{(b)} \texttt{[ENT]} United\_States \texttt{[CON]} support, increase, developed, investments, multinational, agreement, capital, \textit{treaty}, \textit{country}, \textit{investor}, \textit{negotiate} \texttt{[CLAIM]} \textit{America is not a very good example of that, but it does seem like we are in the process of stabilizing.}
        \hlblue{(c)} \texttt{[ENT]} United\_States \texttt{[CON]} hurt, rebound, irony, economy, health, \textit{editor}, \textit{street}, \textit{wall} \texttt{[CLAIM]} \textit{America is not a very good example of that, but it does seem like we are in the process of stabilizing.}
        \hlgreen{(d)} \texttt{[CON]}  tear, national, barrier, capital, flight, lead, kind, starvation, key, essential, tenet, trade, policy
        \\
        \textsc{DYPLOC} & \hlblue{To the Editor: The irony is apparent: the health of the American economy may end up hurting the world, as you point out in your Dec. 9 editorial ``Rebounding Wall Street.''} 
        \hlred{Multinational agreements like the Investment and Important Trade Treaty negotiated in Paris, with the full support of the United States, will increase capital mobility, pointing investors to developed nations and smaller countries with the throat of the global economy. }
        \hlyellow{In addition to limiting rights imposed by environmental restrictions on foreign investors, including ownership of natural resources and media, the treaty will have a cumulative effect on places like traditional national parks and tourist attractions.} 
        \hlgreen{Tearing down national barriers to capital flight will lead to the kind of starvation that is a key essential tenet of trade policy.} \\
        
        \textsc{Seq2seqFull} &  To the Editor: The irony is apparent: the health of the American economy will not end up hurting the world as you point out (editorial,
Dec. 17). The rebound from Wall Street has torn down national barriers to capital flight and led to a kind of starvation. A key essential tenet of
American trade policy is the Multinational Agreement on Investment. This important treaty was negotiated in Paris with the full support of the
United States. It increased the mobility of investors from developed nations to smaller countries, and it has helped keep the United States out of the
world's throat. In addition, it limits rights of nations to impose environmental restrictions on foreign investment, including ownership of natural
resources and media rights.   \\
        
        \midrule
        \end{tabular}
        
        \begin{tabular}{p{20mm} p{140mm}}
       
        \textit{Title} &  Why Save Risky Hedge Fund?  \\
        \midrule
        
        \textit{Content Items} & 
        \hlyellow{(a)} \texttt{[ENT]} Chairman \texttt{[CON]} trust, vice, \textit{international}
        \hlred{(b)} \texttt{[ENT]} Box\_office\_bomb  \texttt{[CON]} bailout, fail, understand, organize, hold, hedge, capital, difficult, federal, \textit{bank}, \textit{governance}, \textit{regulatory}, \textit{term}
        \hlblue{(c)} \texttt{[CON]}  advocate, force, assert, efficiency, free, consistently, principle, \textit{global}, \textit{market}, \textit{community},  \textit{country}
        \hlgreen{(d)} \texttt{[CON]} bailout, arrange, federal, \textit{regulator}, \textit{market}
        \\
        
        \textsc{DYPLOC} & \hlred{To the Editor: Re ``Funds Big and Fail, Fed Assists Bailout'' (front page, Feb. 13): It is difficult to understand why the Federal Reserve Bank of New York should not organize a privately held hedge fund for long-term investment.} 
        \hlgreen{Federal regulators arranged the bailout because the market didn't work.} 
        \hlblue{This country has asserted the efficiency of free-market forces and has consistently advocated free-market principles to the global community.}
        \hlyellow{MICHAEL J. BERMAN New York, Feb. 13, 1999 The writer is vice chairman of fiduciary trust at Prudential International.} \\
        
        \textsc{Seq2seqFull} & To the Editor: Re ``Funds Big and Fail, Fed Assists in Bailout'' (front page, Dec. 17): It is difficult to understand why the Federal Reserve Bank of New York should not organize a privately held hedge fund to manage long-term capital. 
        The Federal Reserve has been the only central bank of central bank management and regulatory governance in the country to assert the efficiency of free-market forces.
        The Fed has consistently advocated the principle that the global community should not allow regulators to arrange the work of hedge funds. 
        MICHAEL J. KAPLAN New York, Dec.'' 17, 1998 The writer is vice chairman of the fiduciary trust at the International Monetary Fund.  \\
        
        \bottomrule
        \end{tabular}

\caption{\fontsize{10}{12}\selectfont 
Sample content items and system output on NYT. System generated concepts and claims are in \textit{italics}.
}
\label{tab:sample-output-nyt}

\end{table*}

\end{document}